# One-Stage Deep Edge Detection Based on Dense-Scale Feature Fusion and Pixel-Level Imbalance Learning

Dawei Dai[1], Chunjie Wang[1], Shuyin Xia[1], Yingge Liu[1], Guoyin Wang[1]
1. College of Computer Science and Technology, Chongqing University of Posts and Telecommunications, Chongqing, China
{daidw@cqupt.edu.cn}

**Abstract**

Edge detection, a basic task in the field of computer vision, is an important preprocessing operation for the recognition and understanding of a visual scene. In conventional models, the edge image generated is ambiguous, and the edge lines are also very thick, which typically necessitates the use of non-maximum suppression (NMS) and morphological thinning operations to generate clear and thin edge images. In this paper, we aim to propose a one-stage neural network model that can generate high-quality edge images without postprocessing. The proposed model adopts a classic encoder-decoder framework in which a pre-trained neural model is used as the encoder and a multi-feature-fusion mechanism that merges the features of each level with each other functions as a learnable decoder. Further, we propose a new loss function that addresses the pixel-level imbalance in the edge image by suppressing the false positive (FP) edge information near the true positive (TP) edge and the false negative (FN) non-edge. The results of experiments conducted on several benchmark datasets indicate that the proposed method achieves state-of-the-art results without using NMS and morphological thinning operations.

# Introduction

Edge and contour detection is a basic task in the field of computer vision and can be used as a basic operation in many complex tasks [1][2][3]. However, there are still many challenges to overcome in this task owing to the complexity of the image background and the existence of edge noise. In recent studies, the deep convolutional neural network (DCNN) model[4] has been widely used in edge detection, and its performance has been greatly improved compared to traditional methods [5][6][7][8][9]. However, edge detection algorithms based on the DCNN still have two very serious problems: (1) Most CNN-based edge detection models [5][6][7] are two-stage, with CNN being first used to obtain a preliminary edge image and some postprocessing operations such as NMS and morphological thinning then being used to obtain clear and correct edge information. (2) The crisp-edge prediction problem, in which the background pixels near the edge of the object are easily misclassified, results in a relatively thick edge in the final prediction.

Our work primarily focuses on the above edge detection problems, which we consider occur as a result of the following main reasons: (1) Numerous up-sampling operations in the decoder cause the feature maps to lack detailed information, resulting in the generated edge becoming very blurred. (2) The pixel-level classification task is very difficult when training a network with a weighted cross-entropy loss, which makes it difficult to accurately distinguish true positive (TP), false positive (FP), true negative (TN), and false negative (FN) pixels information; this causes thick and blurred edge images to be generated. (3) The high-level features can describe the semantic information to a high degree but weaken the detail description, whereas the low-level features have a high degree of detail and location features, but the semantic information is weak. Current feature-fusion methods cannot fully utilize the complementarity of high-level and low-level features.

In this paper, we aim to propose a one-stage model that generates high-quality edge images without post-processing operations. The proposed model still adopts a basic encoder-decoder framework, in which a pre-trained model[10][11] is used as the encoder and the features of each scale in the encoder are concatenated into the corresponding scale layer of the decoder, which provides more detail information. Furthermore, to generate clear edges, we propose a new loss function that is generalized on the Dice coefficient[12] to control FP and FN. In terms of evaluation, LPCB[13] reported that the direct evaluation of the edge image without NMS is very simple and effective. Consequently, in this study, we focused on evaluating the edge images obtained by models without NMS and morphological thinning operations. Specifically, we conducted extensive experiments to evaluate the advantages of our proposed method over previous edge contour detection methods, some instances are shown in Fig. 1. Our contributions can be summarized as follows:

(1) We propose a method that uses a dense-scale feature-fusion mechanism to supply the details that loosed in up-sampling operations and pixel-level imbalance learning to balance FP and FN pixels, and further generate clear and thin edges without NMS and morphological thinning operations.

(2) Our proposed method outperforms previous state-of-the-art methods without using post-processing operations on benchmark datasets.

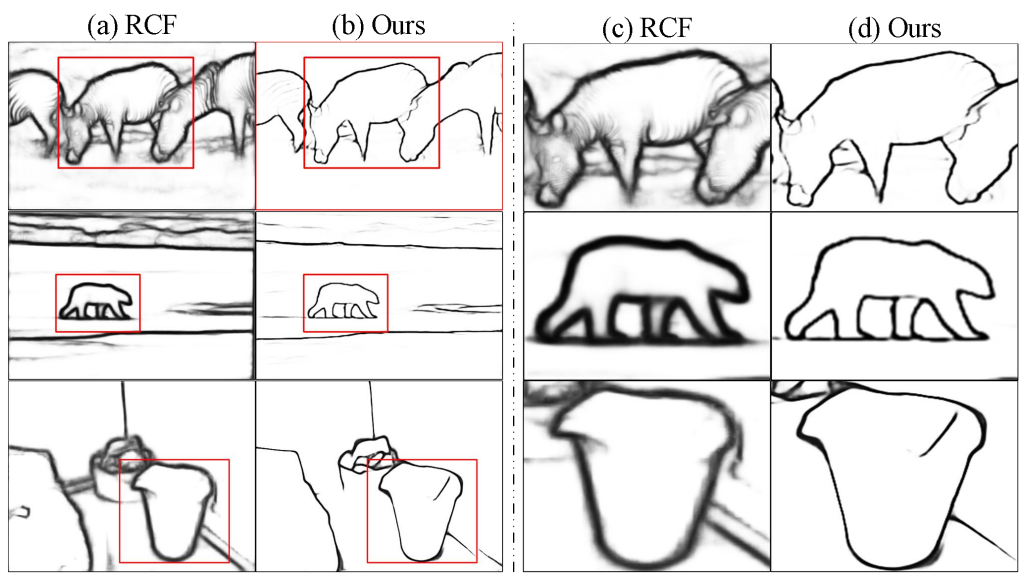

**Figure 1**. Visualization of edge images from RCF and our proposed without post-processing (NMS and morphological thinning). (a) RCF results and (b) results of our proposed. (c, d) Corresponding zoomed images of the (c) RCF results and (d) results of our proposed.

# Related work

**Traditional methods**

Traditional edge detection algorithms generally focus on the change in the gray value of the edge area and that in the frequency domain to obtain the edge contour of the image. For example, the Roberts operator [14] calculates the first derivative of the gray value of the image and defines the pixel at the maximum value as the edge point of the image. The Sobel operator [15] performs convolution operations with pixels in the image using two convolution templates, horizontal and vertical, and regards pixels with gradients greater than a certain threshold as edge points. The Canny operator[16] uses image smoothing, a Gaussian function first-order operator, NMS, a dual threshold algorithm, and boundary connectivity to locate edges. Recent methods have focused on fusing grayscale, gradient, texture, and other features for edge detection, such as pixel boundary (Pb) [17], global Pb (gPb) [18], and statistical edge (SE)[19]. Martin et al.[17] used multiple local cues from luminance (BG), color (CG), and texture (TG) channels as inputs to a regression classifier to predict boundary probabilities. Arbelaez et al. [18] used normalized cuts to extract contours from global information by combining local cues, such as brightness, contrast, and localization, into a globalized framework. Dollar et al.[19] used the structured learning of stochastic decision trees to predict local edge masks, thereby learning accurate edge information.

**Combining CNN and traditional methods**

Ganin et al. [20] proposed the N4-fields method, which extracts the features of each small block using a CNN and then uses the nearest neighbor search on the output of the highest layer of the network to find the edges. On the basis of N4-fields, Bertasius et al. [21] proposed a top-down network called Deep Edge, in which a Canny detector is first used to obtain candidate contour points, then multi-scale features centered on these points are extracted, and finally the scale features are classified and regressed to obtain the probability of edge points. Shen et al. [22] proposed a CNN-based deep contour model that first performs patch processing on the pre-extracted contour information, then performs K-means clustering on small patches, and then uses a CNN to learn the features of each category. However, these methods use the CNN solely as a feature extractor for local detection and do not fully exploit its capabilities.

**Deep Learning-based methods**

Xie et al.[5] proposed an image-to-image network called HED that achieves improved edge detection accuracy using features from convolutional layers. Based on HED, Liu et al.[6] fully utilized the information of all layers, proposed an RCF model, and designed a new loss function that ignores the calculation of controversial edge points to achieve human-level edge detection capability. Yang et al.[23] proposed an encoder-decoder model for the problem of small datasets in edge detection, in which they exclusively optimize the parameters of the decoder to ensure the model's generalization. He et al.[7] proposed a bidirectional cascaded network (BDCN) for multi-scale problems and introduced a scale enhancement module similar to the atrous spatial pyramid pooling (ASSP) structure [24], which enables the network to learn multi-scale representations of different layers. Le et al. [25] proposed the REDN model, which iteratively refines the edges using a recursive encoder-decoder model with skip connections, as well as simple and effective Gaussian blurring-based data augmentation. Inspired by HED and the Xception network, Xavier et al. [26] proposed a robust CNN structure, called DexiNed, to improve the generalization of edge detection.

Wang et al. [27] proposed the CED method for the low spatial resolution of high-level convolution and the similar response of adjacent pixels. Deng et al. [13] theorized that the height imbalance between edge pixels and nonedge pixels is one of the reasons for thicker edges, and consequently proposed the LPCB model. They inserted the ResNeXt module[28] into the encoder–decoder model, and introduced Dice coefficients[12], allowing the CNN to produce sharp boundaries without postprocessing (NMS). Deng et al.[29] proposed a DSCD method for generating high-quality edge information. Inspired by SSIM [30], they proposed a new loss function and added a hypermodule to the model to generate edge features with better quality. Huan et al.[8] proposed the context-aware tracing strategy (CATS) for the mixing phenomenon in a CNN, decomposing boundary features to solve the localization ambiguity problem. Inspired by a previous study [31], Cao et al.[9] proposed an adaptive weighted loss function and a new network DRNet, to stack multiple refinement modules and obtain richer feature representation, thereby achieving crisp boundary information.

Our approach is inspired by Huan et al. [8] and the DSCD method by Deng et al. [29], who both adopted an end-to-end architecture to learn sharp and thin boundaries. Considering the LPCB method by Deng et al.[13], which introduced the Dice coefficient to obtain thin edges, our method is also motivated by the Tversky coefficient [32] to balance the relationship between FP and FN pixels, thereby further improving performance.

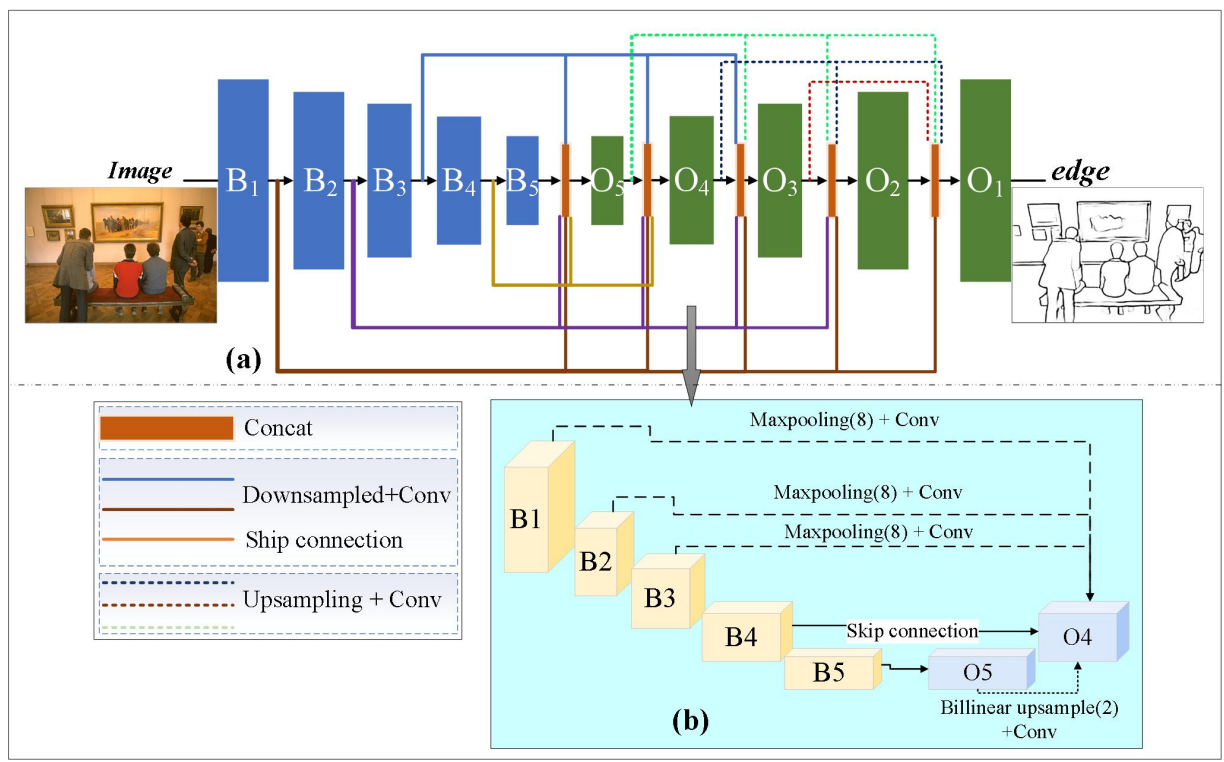

**Figure 2**. Model structure diagram.

# Methodology

**Network architecture**

Our method adopts the convolutional encoder–decoder framework, which is widely used in edge detection. Any pre-trained CNN model can be used as the encoder, and the decoder is implemented by a multilayer feature fuser, which fully exploits more location and detail information in low-level edge graphs and more robust semantic information in high-level features such that the model can learn clearer edges, as shown in Fig.2(a).

$$o4 = C([C(D(B_k)_{k=1}^{i-1}), C(B_i), C(E(o_k)_{k=i+1}^{N})]), \; i=1, ..., N-1 \qquad (1)$$

The proposed multi-feature skip connection not only facilitates interconnection between the encoder and decoder, but also has a connection inside the decoder. This model uses the last layer of each module as the output layer of the encoder, which is defined as [$b_1, b_2,...,b_5$]. For the decoder part, we adopt a multi-layer feature-fusion mechanism. The different output layers, $b_i$, have feature maps of various sizes that can be updated by conventional methods. Here, we take $o_4$ as an example to explain the construction of our decoder part, as shown in Fig.2(b) and Eq.(1), where the function C(.) represents the convolution + batch normalization + ReLU activation operations. D(.) and E(.) denote the up-sampling and down-sampling operations, respectively, and **[.]** denotes concatenation. The feature maps in B4 are fused directly into O4. From the lower layer of the encoder, the max pooling operation is first applied to the feature maps in B1, B2, and B3 and then they are fused into O4. Subsequently, O5 from the decoder part is obtained by bilinear interpolation. To keep the channels unchanged, we adopt a convolutional operation to further unify the number of channels. To better fuse low-level details and spatial features with high-level semantic features, we further perform feature fusion on five scale feature maps.

**Pixel-level Imbalance Learning**

**Weight cross-entropy.** Edge image generation can be regarded as a pixel-level classification task (edge and nonedge pixels). It is extremely unbalanced in terms of the number of edges and nonedge pixels. Weighted cross-entropy (see Eq. (2), where $G_i$ is the label and $P_i$ is the edge prediction value) was adopted to address such a classification task and rectify the imbalance between edge and nonedge pixels, which can effectively supervise the network to distinguish edge and nonedge information. However, the **"thickness"** problem still remains. Although weighted cross-entropy can effectively calculate pixel-level differences, it cannot sufficiently distinguish FP and FN pixel values adjacent to TP edge pixels.

$$f_{ce}(G, P) = -W_i \sum_{i}^{N}(G_i \log p_i + (1-G_i)\log(1-p_i)) \qquad (2)$$

**Tversky-based loss.** Our aim is to generate high-quality edge maps using a CNN without post-processing (NMS and morphological thinning). To this end, we introduce the Dice coefficient into the edge detection and apply softmax along each voxel to generate the loss. We assume that P and G indicate the set of predicted and ground-truth binary labels, respectively. The Dice similarity coefficient D between two binary volumes is defined as Eq.(3).

$$D(P, G) = \frac{2|PG|}{|P|+|G|} \tag{3}$$

We consider that FP and FN differ in the predicted edge map, and, thus, we need to set different weights for the FP and FN pixels. Here, we propose a loss layer based on the Tversky index[33]. The Tversky index defined as Eq.(4), where α and β control the penalties for the FPs and FNs, respectively. We also consider the FN and FP pixel problems as a whole, based on the generalized loss function of the Tversky exponent. For the edge detection, we design our Tversky-based loss function, F1, which is defined as Eq. (5):

$$S(P,G;\alpha,\beta) = \frac{|PG|}{|PG|+\alpha|P/G|+\beta|G/P|} \tag{4}$$

$$f_1 = -\log\left(\frac{\sum_i^N (p_i g_i) + C}{\sum_i^N (p_i g_i) + \alpha((1-p_i)g_i) + (1-\alpha)(p_i(1-g_i)) + C}\right) \tag{5}$$

Where $p_i, g_i$ denote the value of $i^{th}$ pixel on the prediction map P and the ground-truth G, respectively, and C=0.000001. Also, $(1-p_i)g_i$ represents FPs, $p_i(1-g_i)$ represents FNs, α and $(1-\alpha)$ control the weights of FPs and FNs, respectively, where α = 0.7, because we consider that false true pixels are more important in edge detection tasks.

We use cross-entropy loss as the second part of our loss function. The cross-entropy loss focuses on pixel-level differences because it is the sum of the distance of every corresponding pixel pair between the prediction and the ground truth. The final total loss function is defined by Eq. (6). Where $f_{bce}$ is the weighted cross-entropy loss function defined above and β = 0.1 is used to balance the effects of the two losses. Compared with the ordinary weighted cross-entropy loss, the proposed loss function performs better in terms of edge thickness.

$$f_{final} = f_1 + \beta f_{bce} \tag{6}$$

## EXPERIMENTS

**Implementation details**
We implemented our method using the PyTorch framework [34], and evaluated the edge detectors on three benchmark datasets: BSDS500[35] , NYUDv2 [36], and Multicue [37]. An A100 GPU was used for the experiments. The model parameters were as follows: global learning rate (1e-4), mini-batch size (10), weight decay (0.0005), and number of epochs (30). We used the Adam optimizer for optimization. For data augmentation, we followed the methods of LPCB and DSCD by cropping and flipping training image–label pairs by randomly rotating (12 angles) the inscribed rectangles.

**Evaluation standard**

The core step of the edge detection evaluation method is binarization of the prediction result into a binary edge map under the specified maximum allowable distance tolerance and matching of the binary edge map with the ground truth of the edge at the pixel level. On completing the previous matching work, parameter curves of precision and recall are plotted to evaluate the performance of the edge detector. The best performance of an edge detector is reported by the F-measure scores with the optimal thresholds at both the dataset scale (ODS) and image scale (OIS). For the BSDS500 and Multicue datasets, the maximum allowed distance is typically set to 0.0075, and for the NYUDv2 dataset, the maximum distance is set to 0.011. In terms of the evaluation, we adopted the same method as CATS [8] and utilized the following two evaluation protocols.

**Standard evaluation (SEval) protocol.** The evaluation protocol used in previous studies was first subjected to standard postprocessing. The postprocessing scheme typically involved NMS to obtain refined edge maps. The postprocessed edge predictions were then matched with the ground truth to calculate the ODS and OIS.

**Crispness-emphasized evaluation (CEval) protocol.** Although refined edges can be obtained using postprocessing schemes, our intention is that edge detectors should directly generate finer and clearer edge information. Therefore, following the CATS[8], DRC[9] approach, we removed the postprocessing method (NMS and morphological thinning) to evaluate the performance of the edge detectors in terms of edge fineness and sharpness.

Table 1. Our proposed model with various subsystems

| Setting / Methods | Encoder | | Architecture | | Loss function | |
|---|---|---|---|---|---|---|
| | Vgg16 | EfficientNetv2 | Simple | Ours | $f_{bce}$ | Ours |
| Vgg16-w | √ | | √ | | √ | |
| Vgg16-w-f | √ | | | √ | √ | |
| Vgg16-w-f-i | √ | | | √ | | √ |
| EfficientNetv2-w | | √ | √ | | √ | |
| EfficientNetv2-w-f | | √ | | √ | √ | |
| **EfficientNetv2-w-f-i** | | √ | | √ | | √ |

**Ablation experiment**

This section presents ablation experiments conducted to evaluate the performance of all subsystems of our proposed, as shown in Table 1. We only used the BSDS training dataset for training and evaluation of the BSDS test set. We present SEval for performance evaluation as well as the CEval criteria. From the results in Table 2, we can note that (1) dense-scale feature fusion can improve the performance of both SEval and CEval by a large margin, see Vgg16-w *vs.* Vgg16-w-f, and EfficientNetv2-w *vs.* EfficientNetv2-w-f; (2) our proposed Tversky-based loss improves the performance of our models without postprocessing operations, see the CEval of EfficientNetv2-w-f-i. These experimental results demonstrate: both the two subsystems that we proposed worked and achieved our expect in the edge detection task.

Table 2. Comparative experiments on the BSDS500 dataset.

| Methods | SEval | | CEval | |
|---|---|---|---|---|
| | ODS | OIS | ODS | OIS |
| Vgg16-w | 0.797 | 0.808 | 0.602 | 0.602 |
| Vgg16-w-f | 0.808 | 0.821 | 0.677 | 0.677 |
| Vgg16-w-f-i | 0.811 | 0.829 | 0.711 | 0.717 |
| EfficientNetv2-w | 0.816 | 0.827 | 0.640 | 0.640 |
| EfficientNetv2-w-f | **0.818** | **0.833** | 0.696 | 0.704 |
| EfficientNetv2-w-f-i | 0.811 | 0.826 | **0.723** | **0.726** |

**Comparison with state-of-the-art methods**

**BSDS500.** We compared our method with non-deep learning algorithms, such as Canny [16], gPb-UCM[18], and SE[19], and recent deep learning-based methods, specifically HED [5], RCF[6], CED[27], DRC[9] and BAN[38] , among others [13][29][39][40]. We also followed the work of other researchers that leveraged additional training data from the PASCAL VOC Context dataset [41]. Table 3 presents the evaluation results in terms of ODS and OIS, and Fig. 5(a) and (b) shows the precision–recall curves of BSDS500 under the Standard evaluation and Crispness-emphasized evaluation .

Table 3. Comparison with some competitors on the BSDS500 dataset.VOC-aug refers to training with extra PASCAL VOC context data.

| Methods | SEval | | CEval | |
|---|---|---|---|---|
| | ODS | OIS | ODS | OIS |
| Canny(1986) | 0.611 | 0.676 | - | - |
| gPb-UCM(2011) | 0.729 | 0.755 | - | - |
| SE(2015) | 0.743 | 0.763 | - | - |
| HED(2017) | 0.788 | 0.808 | 0.576 | 0.591 |
| RCF(2018) | 0.799 | 0.815 | 0.596 | 0.612 |
| CED(2017) | 0.803 | 0.820 | 0.642 | 0.656 |
| LPCB(2018) | 0.800 | 0.816 | 0.693 | 0.656 |
| BDCN(2019) | 0.807 | 0.822 | 0.634 | 0.650 |
| CATS-RCF(2021) | 0.805 | 0.822 | 0.705 | 0.716 |
| DRC(2021) | 0.802 | 0.818 | 0.697 | 0.705 |
| BAN(2021) | 0.810 | 0.827 | - | - |
| Ours-VGG16 | 0.811 | 0.829 | 0.711 | 0.717 |
| Ours-EfficientNetv2 | 0.812 | 0.827 | 0.723 | 0.726 |
| Ours-EfficientNetv2-VOC-aug | **0.823** | **0.836** | **0.745** | **0.749** |

In Table 3, we can see that our proposed method achieves the current state-of-the-art in ODS and OIS on CEval when trained only with the BSDS500 dataset. The corresponding ODS OIS values are 0.711 and 0.717, respectively. When EfficientNetv2 is used as our backbone, our effect is improved to ODS = 0.812 and OIS = 0.827, reaching a very high level. In the standard evaluation, our model also reached the current state-of-the-art performance. By training with the additional dataset, our model achieves significant improvements. Among them, under the Crispness-emphasized evaluation, ODS f-score from 0.723 to 0.745, and under the Standard evaluation, ODS f-score from 0.812 to 0.823. Fig. 5(a) shows the Standard evaluation results. The performance of the human eye in edge detection is known to have a value of 0.803 ODS F-measure. Our method obtained better results than the average human performance. Our results are also better than the current state-of-the-art (SOTA) results.

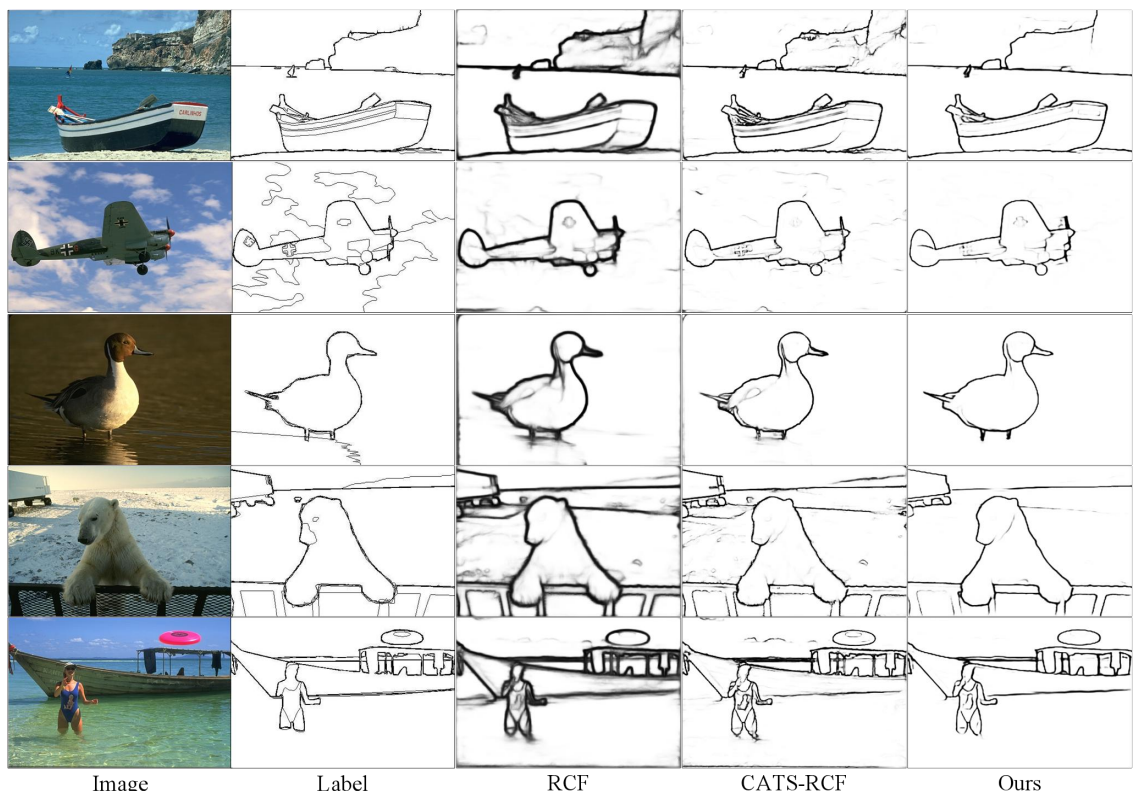

Figure 3. Comparison between state-of-the-art (SOTA) in deep edge detection techniques and our method on the BSDS500 dataset.

The visualization results for the BSDS500 dataset are shown in Fig. 3. In the given examples, our proposed method is effective in suppressing FP pixels near real edge pixels and FN pixels that cause blurring and greatly improves the clarity and thinness of edge images while ensuring accurate edge localization. We can see that our method solves the thick and blurry phenomenon commonly observed in edge detection, producing sharper and thinner edges. Compared with the current research method, CATS, which refines edge detection, the results of our method are better.

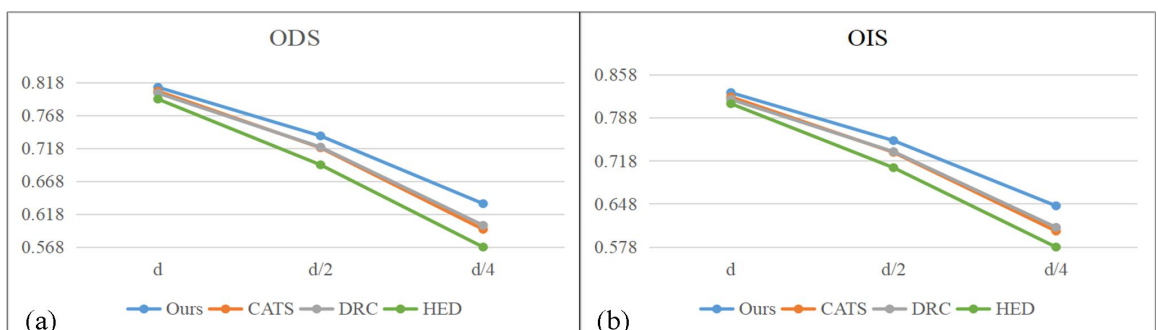

Figure.4 Comparison results for "crisp" edges. The performance (ODS and OIS) is presented as a function of the maximal permissible distance d.

In addition, we followed the CED[27] practice, and further benchmarked the "crispness" of the edges. We evaluated the results by varying the matching distance *d*, and also evaluated our proposed method on the following settings of *d*: $d_0$, $d_0/2$ and $d_0/4$, where $d_0 = 4.3$ pixels (corresponding to the parameter maxDist = 0.0075 for BSDS500) after NMS and morphological thinning. Fig.4 compares our method with HED, CATS, and DRC. The performance of all the methods degrades as *d* decreases. The drop in the HED is evident, and the gap between the other detectors becomes smaller. Conversely, as d decreases, the gap between our method and the other methods increases. In fact, the ODS gap between our method and the previous SOTA level increased from 0.5% to 5.5%, and the OIS gap increased from 0.8% to 5.55%. Meanwhile, our method achieves ODS = 0.634 under the setting of $d_0/4$, which exceeds the human level (0.625) and outperforms the others by a large margin. Thus, it is clear that the proposed method generates crisp-edge maps.

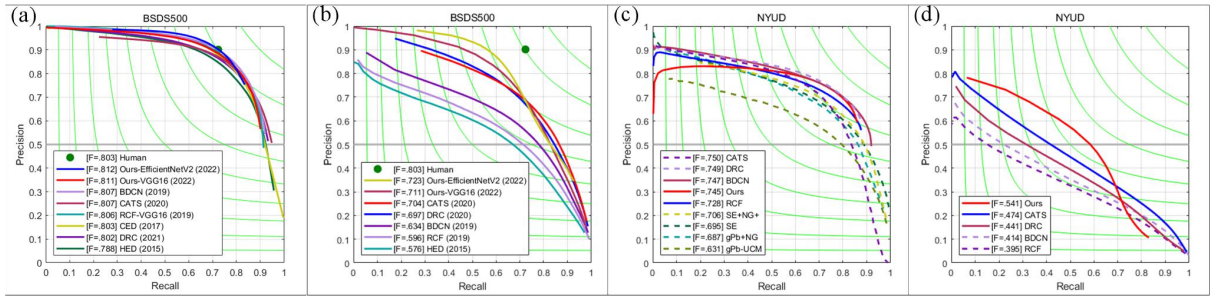

Figure 5. The precision-recall curves of our model and the compared other models on BSDS500 and NYUDv2 datasets. (a) and (c) depict the results under the standard evaluation, while (b) and (d) is the results of the crispness-emphasized evaluation.

**NYUD.** We followed previous work [6][9][13] by independently experimenting on the RGB and HHA data provided in the NYUDv2 dataset and fusing the predicted images based on RGB and HHA to generate merged edge predictions. We compared our proposed method with well-established deep edge detectors. The precision–recall curves are shown in Fig. 4(c). It is clear that, compared with the current SOTA deep edge detectors, our model achieves SOTA performance. A quantitative comparison is presented in Table 4.

In Table 4, we compare our model with the deep edge detectors HED, RCF, LPCB, BDCN, CATS, and DRC. Based on CEval in Table 4 and Fig. 4(d), it is clear that our results are superior to those of the current SOTA. On "Crispness" boundary detection, our method achieves better performance, where ODS = 0.541, OIS = 0.547. Compared with the current edge detectors CATS and DRC for "Crispness" of edge, our method achieves better results on the NYUDv2 dataset. In evaluations that emphasize crispness, our method brings about 14.1% and 12.1% improvements in the ODS and OIS scores, respectively, compared to the CATS method. Fig. 6 shows the qualitative results for the NYUDv2 dataset, where it can be seen that the edge maps generated by our method are clearer and thinner than those of the other models, indicating that our method is more robust in complex environments.

Table 4. Comparison results on the NYUDv2 dataset. SEval and CEval denote the standard evaluation and the crispness-emphasized evaluation, respectively.

| Methods | Data | SEval | | CEval | |
|---|---|---|---|---|---|
| | | ODS | OIS | ODS | OIS |
| HED | RGB | 0.722 | 0.737 | 0.387 | 0.404 |
| | HHA | 0.691 | 0.704 | 0.335 | 0.350 |
| | RGB-HHA | 0.746 | 0.764 | 0.368 | 0.384 |
| RCF | RGB | 0.745 | 0.759 | 0.395 | 0.412 |
| | HHA | 0.701 | 0.712 | 0.333 | 0.348 |
| | RGB-HHA | 0.764 | 0.778 | 0.374 | 0.397 |
| LPCB | RGB | 0.739 | 0.754 | - | - |
| | HHA | 0.707 | 0.719 | - | - |
| | RGB-HHA | 0.762 | 0.778 | - | - |
| BDCN | RGB | 0.728 | 0.762 | 0.414 | 0.439 |
| | HHA | 0.704 | 0.716 | 0.347 | 0.367 |
| | RGB-HHA | 0.766 | 0.779 | 0.375 | 0.392 |
| CATS-RCF | RGB | **0.752** | **0.765** | 0.474 | 0.488 |
| | HHA | **0.710** | **0.721** | 0.433 | 0.445 |
| | RGB-HHA | **0.768** | **0.782** | 0.439 | 0.452 |
| DRC | RGB | 0.749 | 0.762 | 0.441 | 0.455 |
| | HHA | 0.711 | 0.722 | 0.370 | 0.382 |
| | RGB-HHA | 0.769 | 0.782 | 0.403 | 0.436 |
| **Ours** | RGB | 0.745 | 0.756 | **0.541** | **0.547** |
| | HHA | 0.708 | 0.718 | **0.517** | **0.523** |
| | RGB-HHA | 0.758 | 0.769 | **0.527** | **0.534** |

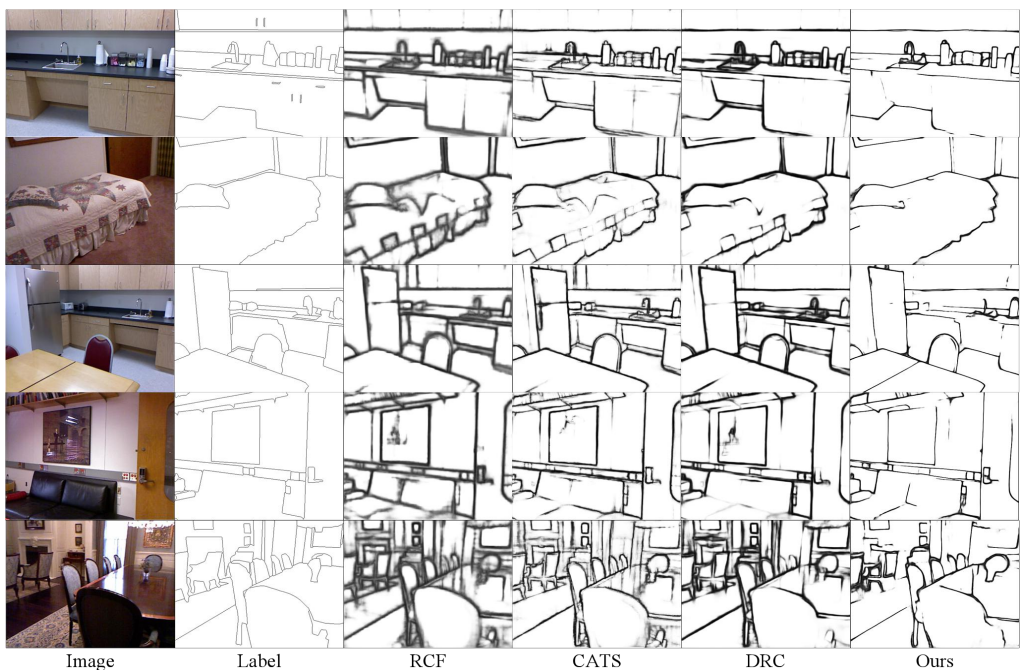

Image    Label    RCF    CATS    DRC    Ours

Figure 6. Comparison between SOTA deep edge detection techniques and our proposed method on the NYUDv2 dataset.

**Multicue dataset.** The Multicue dataset labels both the edge and boundary pixels in each annotated image. Compared with the images in the BSDS500 dataset (321 × 481), the Multicue dataset images had a higher resolution (1280 × 720). We followed the pre-processing steps of RCF and DSCD, first randomly performing random horizontal flips, rotations (90, 180, and 270 degrees), and scaling (75% and 125%) of image–label pairs. During training, we randomly cropped 500 × 500 patches from the original image, and the training hyperparameters were set identically to those of the BSDS500 dataset. All methods were trained three times on boundary and edge data, and the average score of three independent experiments was taken as the final result. The results are presented in Table 5.

Table 5. Comparison results on the Multicue dataset

| Method | ODS | OIS |
|---|---|---|
| Human-Boundary | 0.760(0.017) | - |
| Multicue-Boundary | 0.720(0.014) | - |
| HED-Boundary | 0.814 (0.011) | 0.822 (0.008) |
| RCF-Boundary | 0.817 (0.004) | 0.825 (0.005) |
| DRC-Boundary | 0.820 (0.006) | 0.820 (0.005) |
| Our-Boundary | **0.825(0.002)** | **0.834(0.003)** |
| Human-Edge | 0.750 (0.024) | - |
| Multicue-Edge | 0.830 (0.002) | - |
| HED-Edge | 0.851 (0.014) | 0.864 (0.011) |
| RCF-Edge | 0.857 (0.004) | 0.862 (0.004) |
| DRC-Edge | 0.859 (0.002) | 0.862 (0.001) |
| Our-Edge | **0.886(0.002)** | **0.888(0.001)** |

VGG16 was used as the backbone network. As can be seen in Table 5, our method achieved better results than HED, RCF, and DRC. For boundary tasks, our method had an F-measure 0.6% and 1.7% higher for ODS and OIS, respectively, than DRC. For edge tasks, our method had a 3.1% higher ODS F-measure than DRC. At the same time, our method achieved a 3.0% higher OIS F-measure than DRC.

# CONCLUSION

In the edge detection task, although current SOTA methods have achieved human-level performance on some benchmark standard datasets, edge detection remains an extremely challenging problem, specifically, pixel imbalance. The edge image generated by the depth-edge detector usually needs to be processed by standard postprocessing (NMS and morphological thinning) to obtain accurate localization. Our proposed one-stage model addresses this pixel imbalance via dense-scale feature fusion and pixel-level imbalance learning, which results in improved performance compared with the SOTA results on three benchmark datasets. However, severe pixel-level imbalance remains a challenge, and so in future work we will continue to focus on learning true pixel-level labels that can account for a small proportion of the pixels in one image.